\documentclass[runningheads]{llncs}
\usepackage{xcolor}
\usepackage{graphicx}
\usepackage{amssymb}
\usepackage{pifont}
\newcommand{\cmark}{\ding{51}}%
\newcommand{\highlight}[1]{\colorbox{blue!10}{#1}}

\usepackage{makecell}
\usepackage{amsmath}
\DeclareMathOperator*{\argmax}{arg\,max}
\usepackage{xcolor}

\usepackage{amsmath}
\definecolor{amazon}{rgb}{0.23,0.48,0.34}
\definecolor{avocado}{rgb}{0.34,0.51,0.01}

\begin{document}
\title{Predicting the Ordering of Characters in Japanese Historical Documents}

%
\author{Alex Lamb\inst{1} \and
Tarin Clanuwat\inst{2} \and
Siyu Han\inst{3} \and
Mikel Bober-Irizar\inst{4} \and
Asanobu Kitamoto\inst{2}}
%
%
\institute{Montreal Institute for Learning Algorithms (MILA) \and
ROIS-DS Center for Open Data in the Humanities, National Institute of Informatics \and German Research Center for Environmental Health \and University of Cambridge}
%
\maketitle              
\begin{abstract}
Japan is a unique country with a distinct cultural heritage, which is reflected in billions of historical documents that have been preserved. However, the change in Japanese writing system in 1900 made these documents inaccessible for the general public. A major research project has been to make these historical documents accessible and understandable.  An increasing amount of research has focused on the character recognition task and the location of characters on image, yet less research has focused on how to predict the sequential ordering of the characters. This is because sequence in classical Japanese is very different from modern Japanese. Ordering characters into a sequence is important for making the document text easily readable and searchable.  Additionally, it is a necessary step for any kind of natural language processing on the data (e.g. machine translation, language modeling, and word embeddings).  We explore a few approaches to the task of predicting the sequential ordering of the characters: one using simple hand-crafted rules, another using hand-crafted rules with adaptive thresholds, and another using a deep recurrent sequence model trained with teacher forcing.  We provide a quantitative and qualitative comparison of these techniques as well as their distinct trade-offs.  Our best-performing system has an accuracy of 98.65\% and has a perfect accuracy on 49\% of the books in our dataset, suggesting that the technique is able to predict the order of the characters well enough for many tasks.  
\end{abstract}

\section{Introduction}

The Kuzushiji writing system was used in Japan from the 10th century to the beginning of the 20th century, but due to the standardization of Japanese language textbooks in 1900, Kuzushiji is now only readable by a small number of specially trained scholars.  This has led to a great deal of interest in using machine learning to analyze these historical Japanese texts \cite{clanuwat2018deep,tian2020kaokore}.  A major goal of this research program has been to develop systems for automatically converting these documents into the modern Japanese writing system.  KuroNet \cite{kuronet2019clanuwat}, the first model that achieved high accuracy in Kuzushiji recognition, shows that it is possible to recognize the characters without using character sequences. This is due to the characteristics of Kuzushiji:  that the characters usually overlap and are hard to separate into individual characters. Therefore, KuroNet outputs its result as an unordered list of characters and center coordinates. KuroNet is useful in order to aid human readers by showing the recognized character next to the character in original images even without predicting character sequences.  

The rapid advance in the field continued with the Kaggle Kuzushiji Recognition competition \cite{clanuwat2019kaggle}, in which the winning approach achieved a substantially better F1 score than KuroNet. KuroNet and other top models from the competition mostly used object detection algorithms, and as a result, they only gave results in pixel coordinates. However, character sequences are necessary for machine translation, cataloguing, and search, which are critical for any research that use Kuzushiji documents. The task of predicting the character ordering is simple for some documents, but creating algorithms that can handle the majority of documents is a very challenging task due to irregular character layouts, which is common in pre-modern Japanese documents.

In this paper we explore the task of automatically converting these coordinate lists into a single character sequence for a page according to a predicted reading order. The end goal is to create an algorithm that can handle any type of document given character pixel coordinates in any order.



\subsection{Challenges in predicting the order of characters in Japanese historical documents}

The Kuzushiji writing system was used in Japan for over a thousand years, but due to the standardization of Japanese language textbook in 1900, Kuzushiji is now only readable by a small group of people. Since Japan has over a billion of historical documents written in Kuzushiji, but very few people can read, using automated tools to make the documents accessible to the general public is the highest goal in Kuzushiji recognition research.

The reading order in classical Japanese is very different from modern Japanese. Most classical Japanese text were written in vertical column which is read from the top-right to the bottom-left. This still applies to modern Japanese. Some short text can be written horizontally. In modern Japanese, the reading order for a horizontal row is from left-to-right, whereas in classical Japanese, it was right to left.

Even though a lot of pre-modern Japanese documents were written in clean and separated columns, this layout is far from universal. The reading style can be differ from document to document due to the layout. Irregular text layouts can make reading order extremely hard to determine. One of the reasons for this difficulty is the "Chirashigaki" writing style. 

\begin{figure}
    \centering
    \includegraphics[width=0.7\linewidth]{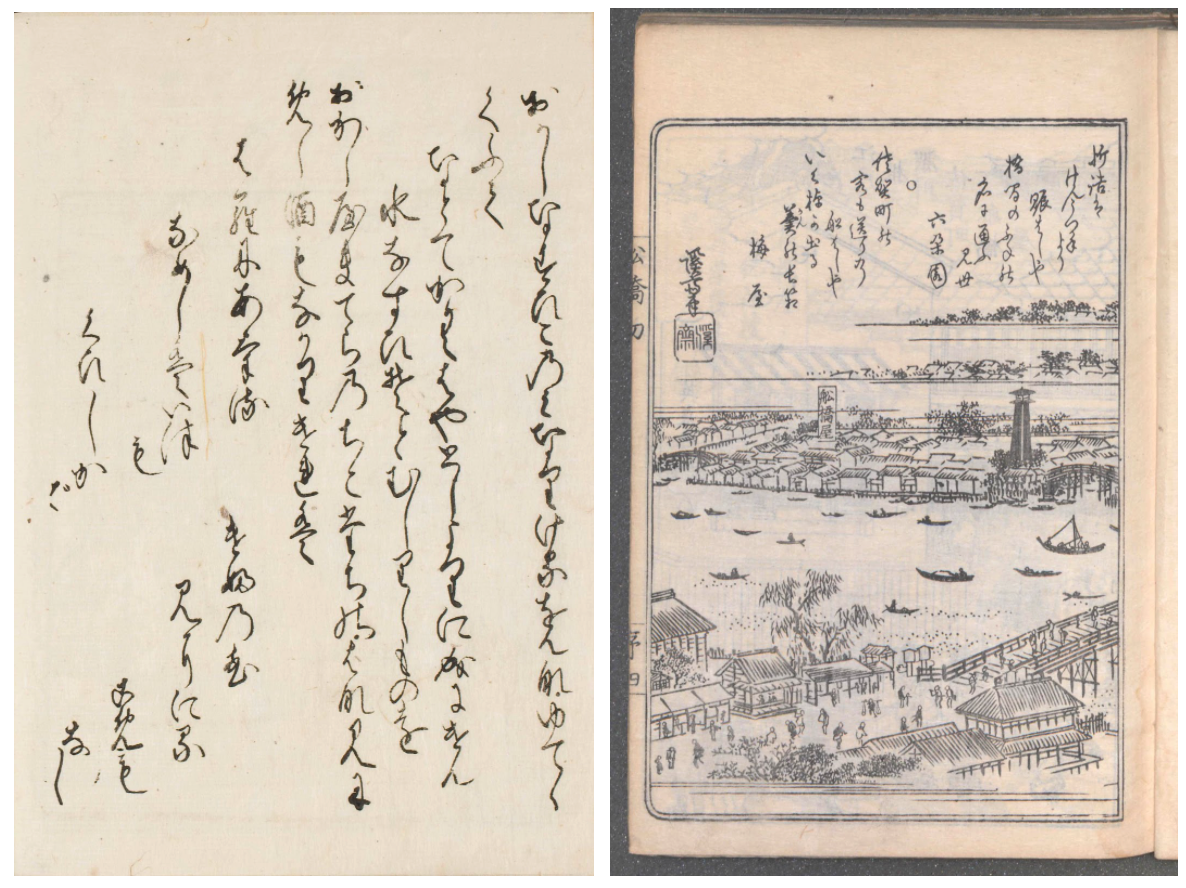}
    \caption{Examples of pages with Chirashigaki layout style from the Kuzushiji dataset.}
    \label{fig:chirashigaki}
\end{figure}

Chirashigaki style was a very common writing style for poems and Hiragana letters. It was popular because of the beauty of the layout (Figure~\ref{fig:chirashigaki}). A writer would be considered skillful if they could use this writing style elegantly. Instead of writing the characters in straight columns, the text was written in curvy and ascending lines. The space between columns was also varied. Chirashigaki was common in handwritten documents, however a lot of woodblock printed documents also tried to imitate the Chirashigaki style. Handwritten documents and books had higher rank than printed books and were considered to be more valuable. Therefore, publishers tried to use some of these styles in printed books.

\begin{figure}
    \centering
    \includegraphics[width=0.8\linewidth]{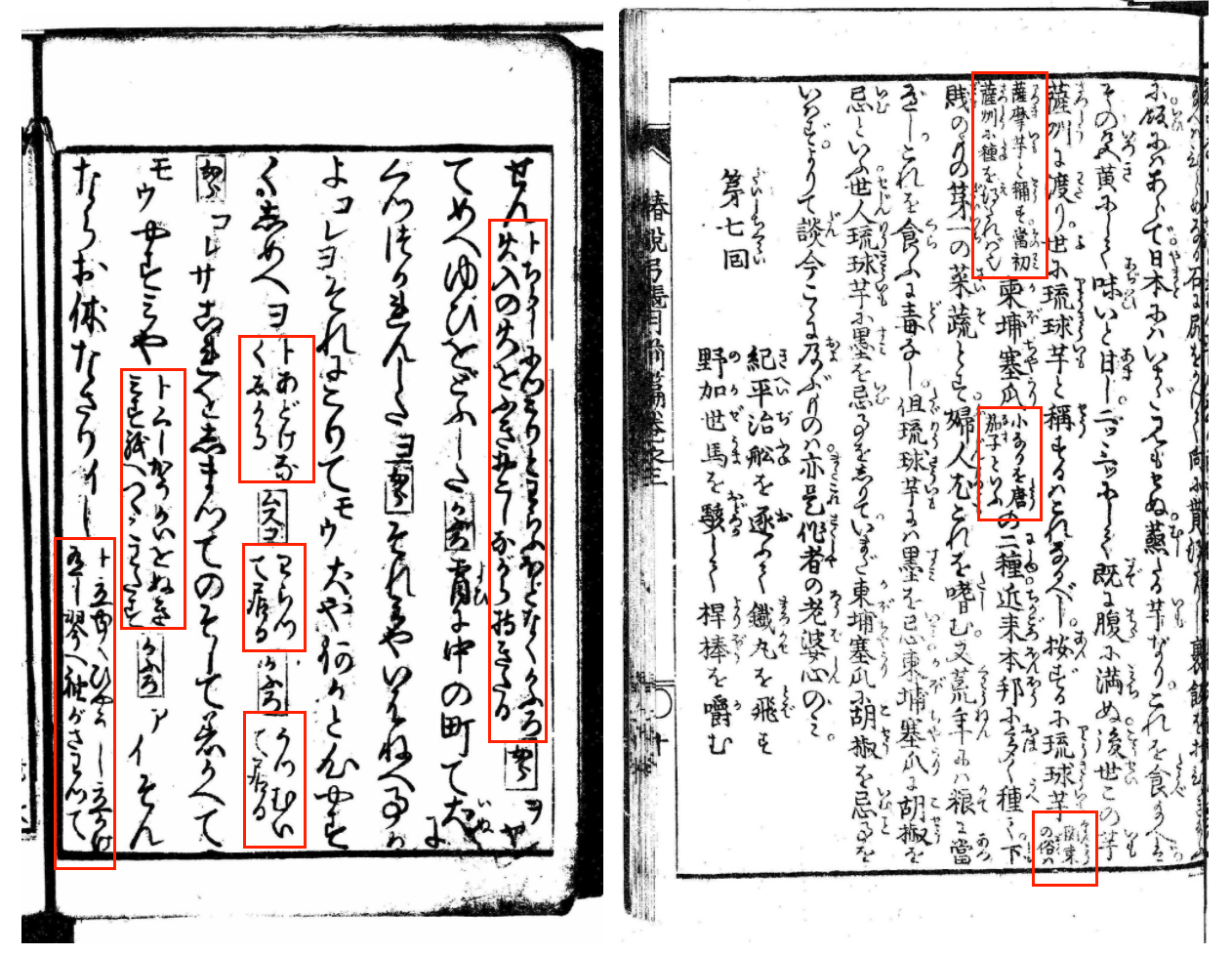}
    \caption{Example of pages with Warichū double-column reading order in the Kuzushiji dataset.}
    \label{fig:warichu}
\end{figure}

Another difficulty in determining the ordering of characters comes from “Warichū” or double smaller columns which, in most cases, are annotations (Figure~\ref{fig:warichu}). Even though small annotations, like those suggesting how to read Kanji characters (Rubi), is omitted in the Kuzushiji dataset. we use in this experiment, Warichū is still considered as the main text. Therefore, it is necessary for the model to determine the ordering of the characters in these double text columns.  

\begin{figure}
    \centering
    \includegraphics[width=0.4\linewidth]{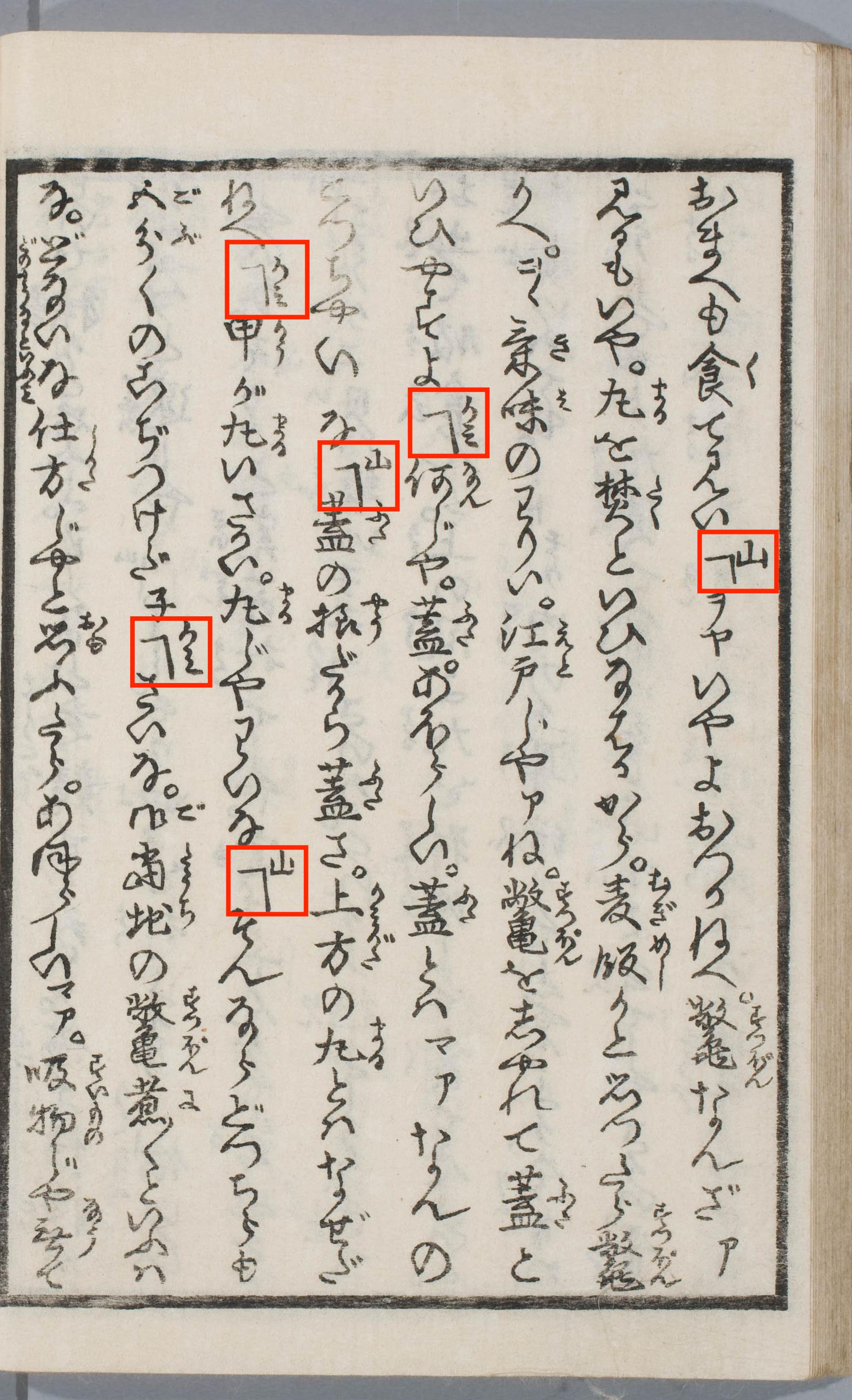}
    \includegraphics[width=0.4\linewidth]{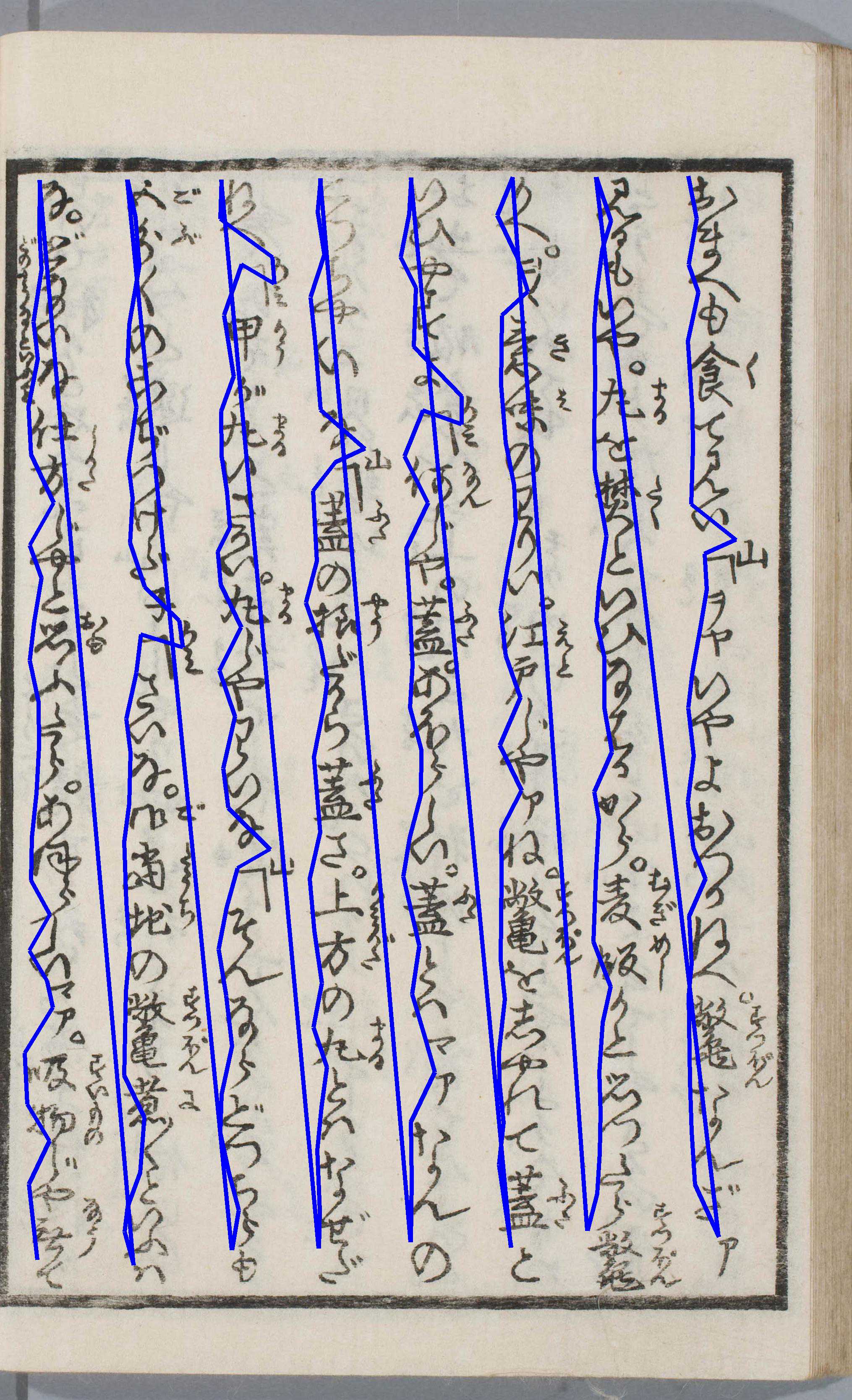}
    \caption{Examples of pages with quotation marks with the identity of the speaker written slightly to the right of the main text column (left) and the reading order ground truth (blue line in right) from the Kuzushiji dataset.}
    \label{fig:quotation}
\end{figure}

Normally classical Japanese documents have limited punctuation, however in the Edo period (17th century - 19th century) we can see that in some story books, there is the use of a symbol that was used as a quotation mark(Figure~\ref{fig:quotation}). Additionally, there is a special symbol which marks the characters that specify which character is speaking in a story. The reading order for these characters would be slightly to the right of the main column.  

In addition to these layout problems, woodblock printed book layouts can be in various style because woodblock were carved as a whole page instead of using movable fonts like in western books. In Japan, there was also movable font printing. However, due to the amount of characters and connectedness between characters, movable font was not popular until the 20th century.  


\subsection{Overview of Approaches}

We propose and evaluate the performance and trade-off of three different methods for predicting the ordering of Kuzushiji characters.  First, we consider a simple rule-based model based on the fact that Japanese is written top-to-bottom, right-to-left, and show that it works on many pages which have this regular layout.  However, the model fails on pages with more complicated but fairly common layouts, such as lines of text which are partly separated into two columns.  We then improve on this with an Adaptive Rules-Based model, in which rule selection is controlled by dynamic thresholds based on the properties of the characters present in the page, and show that this improves performance on many pages with irregular layouts.  To try to automate this process of discovering rules and handling edge cases, we trained a deep recurrent neural network to directly predict the sequential ordering of the characters.  


\section{Related Work}

\paragraph{Kuzushiji Character Recognition: }

The task of Kuzushiji character recognition in KuroNet model consists of determining characters and their locations in historical documents.  Recent research has seen significant progress on the kuzushiji optical character recognition task.  The KuroNet model \cite{kuronet2019clanuwat,lamb2020kuronet,clanuwat2018end} uses a U-Net on top of an entire page of text to directly predict the characters in the page and their positions.  Datasets were also released for classifying images of kuzushiji characters in isolation \cite{clanuwat2018deep}.  A Kaggle competition was also hosted on Kuzushiji character recognition, which saw further improvement over the KuroNet model \cite{clanuwat2019kaggle}.  

Earlier work on this problem focused on recognition using character sequences as training data \cite{nguyen2017kuzu}, but this approach has run into a few difficulties and couldn't achieve high enough accuracy that make the recognition result useful.  Firstly, it is not an end-to-end approach, as it requires the identification and localization of columns of text that can be run through the model. However, when Kuzushiji documents are written in complex layouts without appropriately sized and easily identified columns, such as figure \ref{fig:warichu}, the algorithm failed to produce meaningful output. Secondly, models using this approach needs a lot of preprocessing which may make the recognition task very slow.  

\paragraph{Predicting Ordering and Layout in Historical Japanese Documents: }

The closest work that tried to determine Kuzushiji reading order was done by \cite{le2019kuzu}. Human reading behavior was defined as people determining the start character of a paragraph/line, with the idea that they move the eyes from the current character/word to the next character/word. They can also determine the end of a line or skip a figure to move to the next line" which maybe true for common text. However, the model for this approach doesn't attempt to solve any of the challenges associated with irregular layouts or challenging Kuzushiji reading orders.

\section{Proposed Methods}
\label{sec:methods}

\subsection{Overview of Techniques}

We present and compare three different techniques for predicting the sequential ordering of characters in historical Japanese documents.

The first model, which is the simplest approach, uses a fixed set of rules for reading through a document.  It begins in the top-right of a document, and reads down from that character, until a distance threshold is exceeded, at which point the document selects the top-right remaining point as the start of the next column.  We refer to this as our \textsc{Simple-Rules} model. The second model is an \textsc{Adaptive-Rules} model which follows the same principle, but the rules for when to end a column of text and search for the start of the next column are not fixed thresholds, but are rather dynamic thresholds which are based on the distribution of character sizes and distances in the page being evaluated.  In both of our rules-based approaches, the characters are first grouped into columns, which are read from right to left, and the characters within a column are red in a top-down ordering.  Note that in both \textsc{Simple-Rules} and \textsc{Adaptive-Rules}, there is no training phase and no requirement for training data (aside from a manual tuning of hyperparameters on the training set).  

The third model is a Deep Autoregressive Sequence Model, in which a recurrent neural network is tasked with predicting the next character position in the sequence.  This approach, which we call \textsc{Deep-AR} is potentially more powerful and able to learn more complex layout patterns, yet has the disadvantage that it requires labeled training data as well as an expensive training phase.  Additionally, it is more computationally expensive to run a deep network during the prediction phase.  When evaluating the methods results, these trade-offs should also be carefully considered (Table~\ref{tab:tradeoffs}).  



\setlength{\tabcolsep}{6pt}{
\begin{table}[h!]
\centering
\caption{A comparison of the properties of our three proposed techniques.}
\begin{tabular}{|l|c|c|c|c}
\hline
Approaches & \textsc{Simple-Rules} & \textsc{Adaptive-Rules} & \textsc{Deep-AR} \\ \hline
Requires Training Data &  &  & \cmark  \\
Explicit Domain Knowledge & \cmark & \cmark  & \\
Handles irregular spacing & & \cmark & \cmark \\
Reads Multi-column &  & \cmark & \cmark	\\   
Reads Irregular Layouts &  &  & \cmark	\\   
Uses Visual Layout Clues & & & \cmark		\\   
\hline
\end{tabular}
\label{tab:tradeoffs}
\end{table}}

\subsection{\textsc{Simple Rule-Based Model}}
Our first algorithm or the simple rule-based model involves scanning through the page, starting with the character that appears in the top-right.  This uses a fixed distance threshold to determine when to terminate reading a column and begin searching for the start of the next column.  On some pages, this produces surprisingly good results.  However, this approach is very vulnerable to any variations in layout and easily produces “false positives” when predicting the starting point for new columns.  This model is the simplest algorithm and is very fast at prediction time, but the model is vulnerable to  pages that have slightly irregular layout.  

\subsection{\textsc{Adaptive Rule-Based Model}}

This approach is similar to the \textsc{Simple-Rules} method, but uses adaptive thresholds based on the position and size of characters in the page.  In the \textsc{Adaptive-Rules} approach, when we consider adding a new character to the end of a column, we test to see if its minimum x-distance to all of the other characters in the column is greater than the average width of all of the characters on the page.  If this condition is not met, the character is not added to the end of column, and the model begins the procedure for searching for the beginning of the next column.  Additionally, the search for the beginning of the next column is adaptive to account for parts of text lines which have multiple columns.  This is done by searching for characters which have bounding boxes which substantially overlap in the x-axis with multiple other characters at the same position in the y-axis (for example, a character above a double column of text will overlap with the bounding boxes of characters in both columns of text).  


\subsection{Deep Autoregressive Character Ordering Model}

The Deep-AR model proceeds in a few stages.  First, we use a dense prediction model to extract features from an image of the entire page.  Then we extract an embedding vectors $\mathbf{h} = h_1, h_2, h_3, ..., h_n$ for all the positions in the image where a character is present (using the ground truth character positions).  In the next step, we use an autoregressive sequence model which runs over these position embeddings in-order, and on each step is tasked with predicting the character at the next position in the sequence.  We frame this prediction task as predicting the position-embedding $h_i$ for the next step in the sequence, which is similar to the pointer networks technique used for learning graph traversals \cite{vinyals2017pointer}.  We also mask the predictions to prevent the model from predicting character positions which it has already predicted.  During the prediction phase, we do not have access to ground-truth character sequences, so we use the models own outputs as predictions to condition on for the previous steps \cite{lamb2016professor,goyal2017actual}.  At the same time, the model may reach the end of the sequence without predicting all characters, so we use a nearest-neighbor heuristic for filling in the characters which the model misses during prediction.

\paragraph{Extracting Position Features: }
We used the Efficient-UNet variant \cite{tan2019efficientnet} of the U-Net architecture \cite{ronneberger2015u} to compute the features.  Our architecture for the U-Net closely follows the KuroNet algorithm \cite{lamb2020kuronet}, which also extracts per-position features from an entire page of kuzushiji text.  
\paragraph{Autoregressive Character Prediction: }
We refer to the predicted-position on each step as $\hat{\mathbf{y}} = \hat{y}_1, \hat{y}_2, ..., \hat{y}_n$ for the $n$ characters in the sequence.  We refer to our ground truth sequence as $\mathbf{y} = y_1, y_2, ..., y_n$.  From our U-Net features, we extract the output hidden state at all positions with characters $\mathbf{h} = h_1, h_1, h_3, ..., h_n$.  Note that while these sequences have the same length by construction, they do not have the same ordering (the ordering of the elements of $\mathbf{h}$ is arbitrary and is not used by the model).  

Our autoregressive prediction model consists of two components: a GRU (which stores information about the previously predicted characters) and an attention mechanism (which can query information about all of the character positions in the example).  

The attention mechanism can be written as $A_{t} = \mathrm{MHA}(q = W_qs_t,\quad k = W_k\mathbf{h},\quad v = W_v\mathbf{h})$, where $\mathrm{MHA}$ is a standard multi-head attention mechanism and ($W_q, W_k, W_v$) are learnable parameters.  Note that because the attention mechanism is permutation invariant, it does rely on any information about the ordering of the elements in $\mathbf{h}$.  We can write one step of the model as $s_{t+1} = \mathrm{GRU}_{\theta}(s_t, [h_t, A_t])$ such that it updates the recurrent state and uses a concatenation of the last position's hidden state (from the U-Net) with the attention result as its input.  We then map this to an output state $o_t = W_o s_t$.  We estimate the probability of the next character in the sequence: 

\begin{align}
p(\hat{y}_t|y_{1..t-1},\mathbf{h}) = \mathrm{softmax}(o_t \mathbf{h}^T)
\end{align}

Thus the output hidden state $o_t$ will select a step j as the predicted next element in the sequence when the dot product between $o_t$ and $h_j$ is large.  The training objective is the usual negative log-likelihood based cross-entropy, where the ground truth position $\hat{y}_t$ is used as the target.  

\paragraph{Prediction and Post-Processing: }
While the model is trained to make one-step-ahead predictions, we are interested in predicting the entire sequence of characters for the page.  In principle, we could do this by sampling from the model one step at a time, while conditioning on the previously sampled positions.  However in practice, we found that it worked better to do greedy sampling, where we take the argmax of the distribution on each step.  Additionally, we mask out elements of the softmax corresponding to positions which the model has already predicted to prevent the model from predicting the same character positions twice.  We run this prediction loop until we reach the end-token.  Since the model may have skipped some characters, we re-insert the skipped character either directly before or after the nearest neighbor predicted-character, depending on whether the skipped character is above or below its nearest neighbor (using the property that the language is written top-to-bottom).  While this use of nearest-neighbors is a heuristic, we found that the model typically only skips a handful of characters, and that this approach usually re-inserts them into a suitable place in the sequence.  

\begin{align}
\tilde{y}_{k} = \argmax_{\hat{y}_k} p(\hat{y}_{k} | \tilde{y}_{1:k-1}, \mathbf{h})
\end{align}

\paragraph{Training Details: }
We trained for 150 epochs with a batch size of 12, and a resolution of 672 x 672.  We used the SGD with momentum as the optimizer with a learning rate of 0.01 and a momentum term of 0.9.  The model, which consists of a U-Net and an RNN with multi-head attention, is trained end-to-end.  


\section{Experiments}

The purpose of our experiments is to study how well our proposed models (\textsc{Rules}, \textsc{Adaptive-Rules}, and \textsc{Deep-AR}) are able to accurately predict the sequential ordering of characters in Japanese historical documents, and to study cases where the models differ in their performance.

\subsection{Data}

The Kuzushiji dataset that we use to train and evaluate KuroNet was created by the National Institute of Japanese Literature (NIJL) and released in 2016\footnote{\url{http://codh.rois.ac.jp/char-shape/book/}}.  It is curated by the Center for Open Data in the Humanities (CODH).  The dataset  has over 684,165 character images and consists of bounding boxes for the location of all of the characters in the text as pixel coordinates along with the sequential ordering of characters on the page.  We uniformly divided the dataset into 3961 training images and 416 validation images, and used this split for all of our experiments.  

For the text conversion experiments, we use the data from Kuzushiji Dataset which was created by the National Institute of Japanese Literature (NIJL) and curated by the Center for Open Data in the Humanities (CODH).  The dataset was also used as the train and test data for Kaggle Kuzushiji Recognition Competition.
 The Kuzushiji dataset consists of 3 main parts. First is document page images. Second is character unicode and pixel coordinates in CSV file. Third is cropped character images from each page. In the pixel coordinates CSV file, there is also a Char ID column which specify reading order by the index of characters.

\subsection{Evaluation Criteria}

We considered several different criteria for evaluating our sequence ordering models.  For all of our criteria, we consider a ground truth ordering (1,2,3,4,5,...,N) as well as a predicted ordering (e.g. 1,3,2,4,5,..., N).  For all of our models, the predicted ordering always has the exact same elements as the ground truth ordering, but potentially in a different order than in the ground truth sequence.  

While the simplest criteria would be to compute a simple accuracy score between these two orderings, this would be too harsh of an evaluation criteria, because the model could have a very low accuracy if just a single character is in the wrong place.  For example, consider a ground truth sequence (1,2,3,4,5) and a predicted sequence (5,1,2,3,4).  Although most of the sequence is predicted correctly, it has a position-wise accuracy of 0\%.  

The first criteria we considered is at edit-distance based accuracy, where we first compute an edit distance between the two ordering sequences.  This edit distance is the minimum number of replacement, insertion, and deletion operations required to make the ground truth ordering and the predicted ordering identical \cite{levenstein1966edit}.  Returning to our previous example of (5,1,2,3,4), we can make it match the ground truth sequence with 2 edit operations by first deleting the 5 and then inserting it at the end of the sequence.  Our edit-distance based accuracy score is normalized based on the length of the sequence $|y|$: 

\begin{align}
    \mathrm{Accuracy}(\hat{y}, y) = 1.0 - \frac{d(y,\hat{y})}{|y|}
\end{align}

Edit-distance may still be too strict as an evaluation metric, especially if we are interested in searching for short sub-strings.  If for example, our model reads the wrong half of the page first, but still reads that half of the page in the correct order, its results could still be useful for information extraction tasks (for example searching for short query strings) but would have very high edit distance from the ground truth sequence.  To get around this, we also considered the use of recall metrics, where we consider all queries from the ground-truth text of a given length and measure what fraction of those queries appear in-place in the predicted ordering.  For example, if the ground truth ordering is (1,2,3,4,5), we can consider the following possible two-character queries: (1,2), (2,3), (3,4), (4,5).  If our predicted ordering were (5,1,2,3,4), then we would match 4/5 of the possible queries, giving a recall of 80\%.  Normally precision-and-recall are predicted together, but in most of our experiments, our predictions are a permutation of the ground-truth ordering (same set of elements), and so we only need to report recall since precision-recall trade-off is the same for all of the models.  

\section{Results and Comparisons}

\begin{figure}
    \centering
    \includegraphics[width=\linewidth]{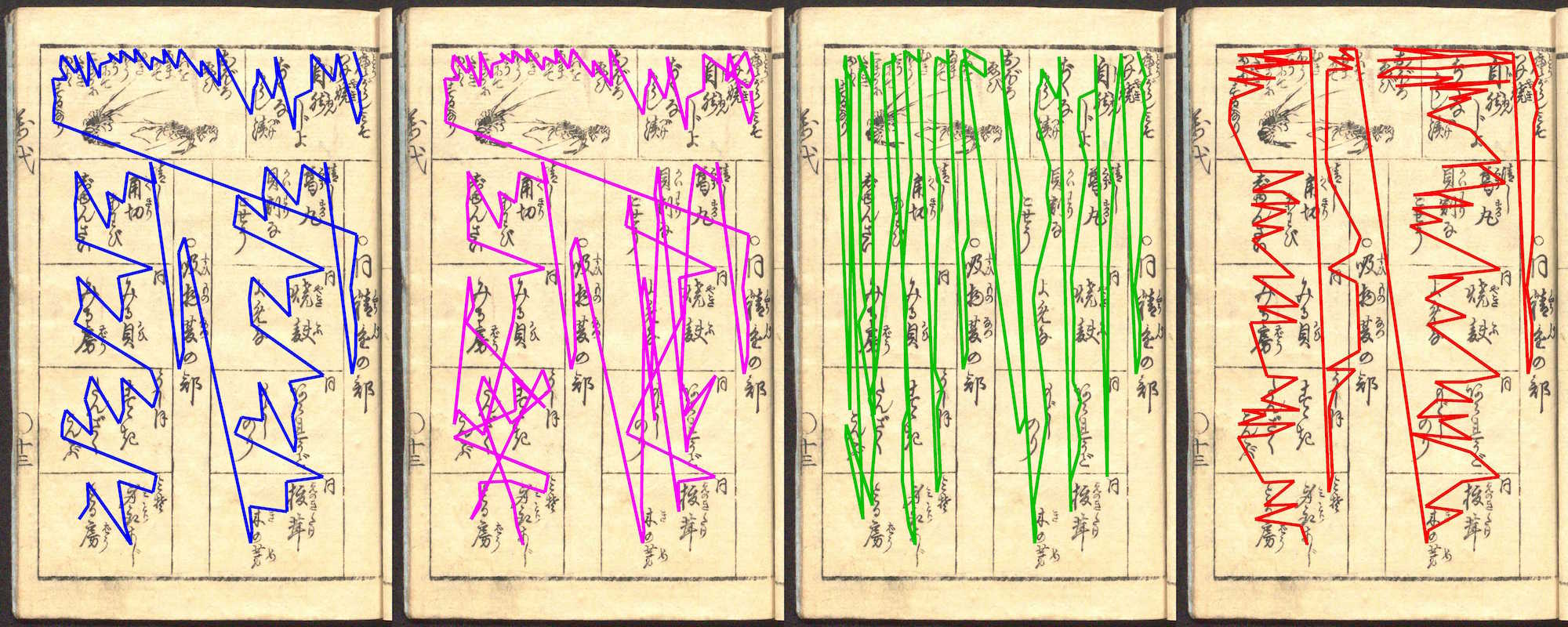}
    \includegraphics[width=\linewidth]{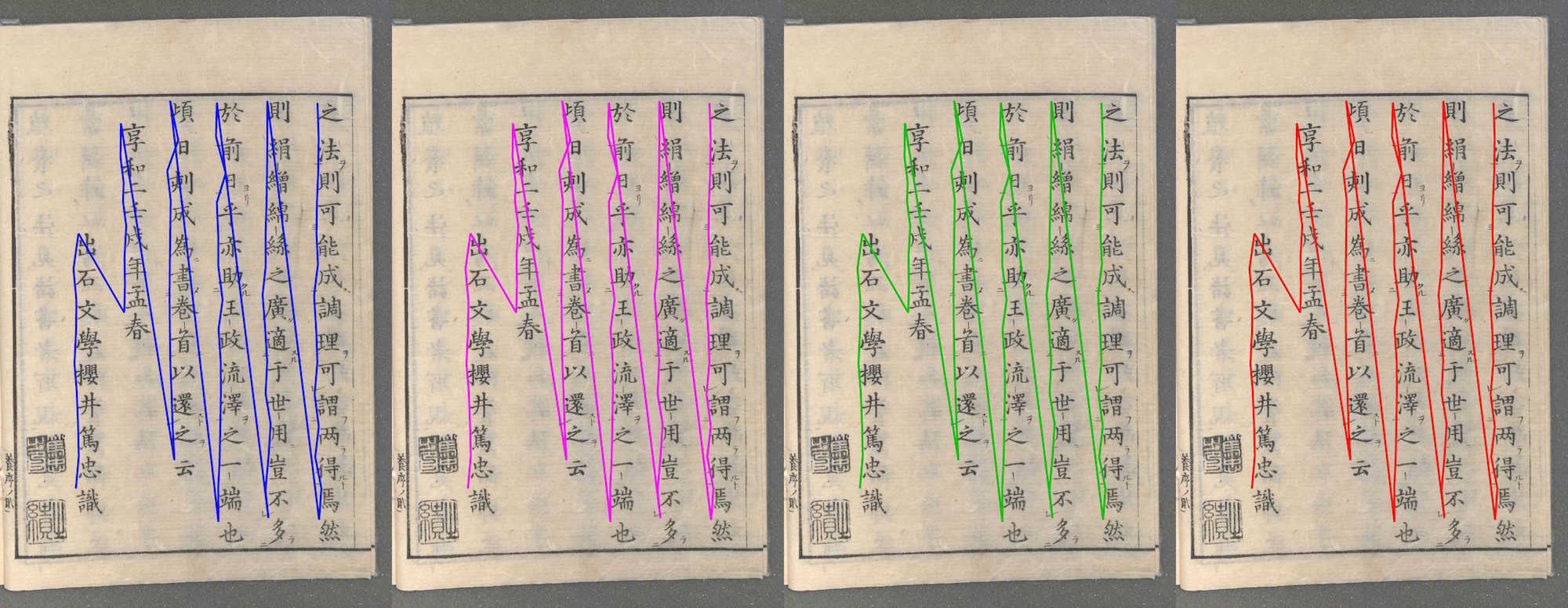}
    \includegraphics[width=\linewidth]{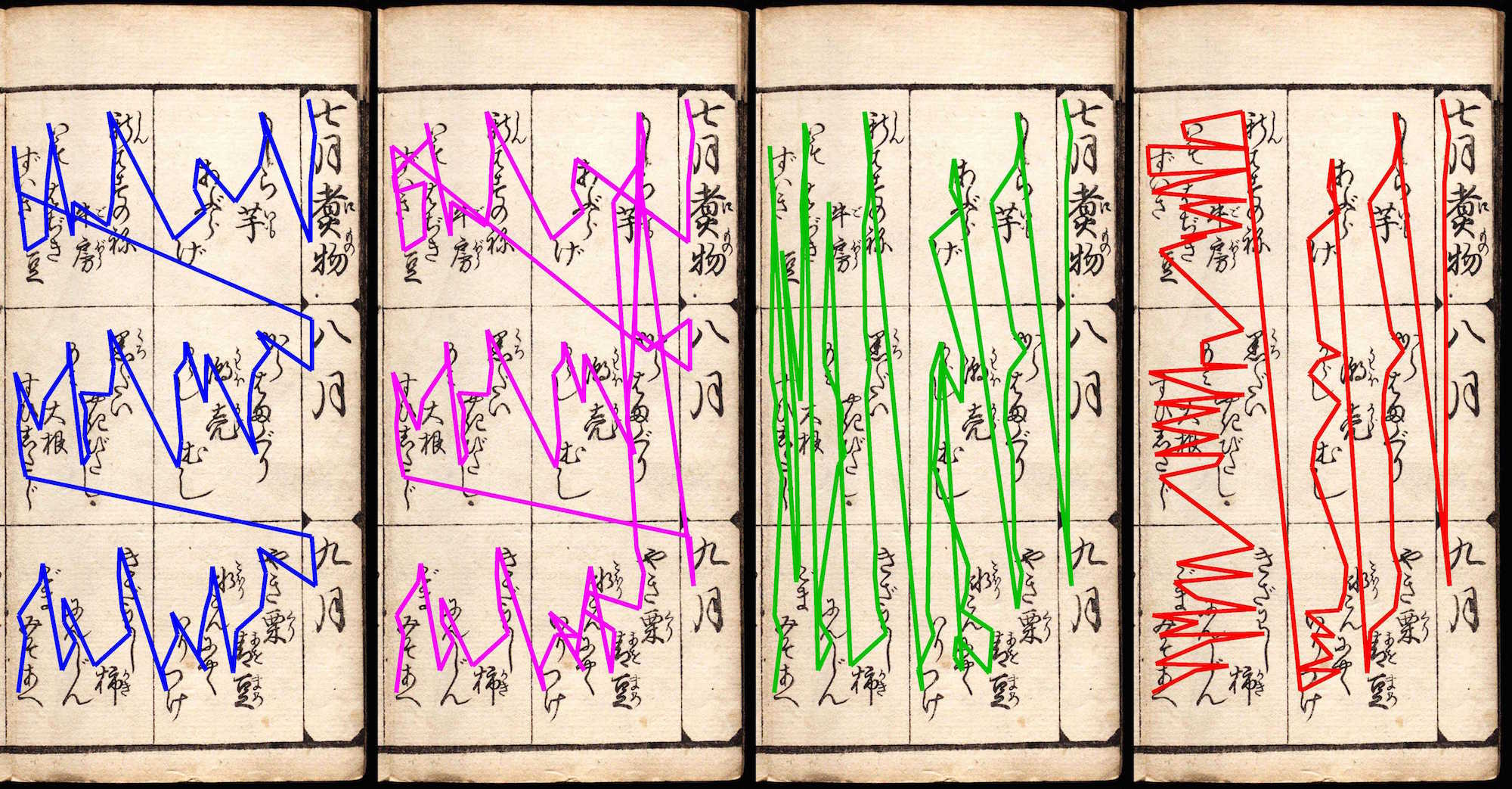}
    \caption{Reading order for three different pages.  \textcolor{blue}{Ground Truth (Blue)}, \textcolor{violet}{\textsc{Deep-AR} (Violet)}, \textcolor{avocado}{\textsc{Adaptive-Rules} (Green)}, \textcolor{red}{\textsc{Rules} (Red)}.  The deep model has the best performance on pages with complex or irregular layouts.  }
    \label{fig:graphic_results}
\end{figure}

We present qualitative and quantitative evaluations on the performance of our three different proposed models.  In Table~\ref{tab:accrate}, we report edit-distance based accuracy and recall performance over characters queries of length 2-21 in Table~\ref{tab:recallrate}.  We also show how recall depends on the character query length in Table~\ref{tab:recallquerylength}.  A particularly notable result is that the performance shows substantial variability between the different books, with a large fraction of the books having perfect performance using all three of our proposed models.  On the other hand, several books were challenging for all of the models.  On the whole, \textsc{Deep-AR} had the strongest performance, with the best performance on 15 out of the 35 books in our dataset, and equally good performance on 14 out of the 35 books, and worse performance on only 6 out of the 35 books.  We also plot examples of reading orders in Figure~\ref{fig:graphic_results}, showing improved handling of irregular layouts with \textsc{Deep-AR} and \textsc{Adaptive-Rules}.  



{\renewcommand{\arraystretch}{1.0}
\setlength{\tabcolsep}{16pt}{
\begin{table}[h!]
\centering
\caption{Accuracy Rate (based on edit-distance) by Book for each Model.}
\begin{tabular}{|l|c|c|c|}
\hline
Book ID& \textsc{Simple-Rules} & \textsc{Adaptive-Rules} & \textsc{Deep-AR} \\ \hline
hnsd00000 & \highlight{99.80} & 99.77 & 99.80 \\
brsk00000 & 85.05 & 99.26 & \highlight{99.39}	\\ 
100249476 & 90.13 & 99.14 & \highlight{100.0}	\\ 
200021851 & 98.86 & \highlight{100.0} &	\highlight{100.0} \\ 
200018243 & \highlight{100.0} & \highlight{100.0} & \highlight{100.0}	\\ 
200021802 & 95.16 & 99.28 &	\highlight{99.79} \\ 
200005798 & 99.88 & 99.76 &	\highlight{100.0} \\ 
200010454 & \highlight{100.0} & \highlight{100.0} &	\highlight{100.0} \\ 
200021660 & 92.8 & 96.31 & \highlight{97.67}	\\ 
200003076 & 98.76 & 99.61 &	\highlight{99.68} \\ 
100249376 & \highlight{100.0} & \highlight{100.0} & \highlight{100.0}	\\ 
200004107 & \highlight{100.0} & 99.32 & \highlight{100.0}	\\ 
200015779 & 85.26 & 93.98 & \highlight{98.74}	\\ 
umgy00000 & 90.67 & 97.25 & \highlight{98.80}	\\ 
100249416 & \highlight{100.0} & 99.21 & \highlight{100.0}	\\ 
200004148 & 79.18 & 96.52 &	\highlight{97.88} \\ 
100249537 & \highlight{100.0} & \highlight{100.0}  & \highlight{100.0}	\\ 
200003803 & \highlight{100.0} & \highlight{100.0} & \highlight{100.0}	\\ 
200021712 & \highlight{100.0} & \highlight{100.0} & \highlight{100.0} \\ 
200021853 & 39.37 & 39.19 &	\highlight{93.53} \\ 
200014740 & 99.86 & 99.39 &	\highlight{99.91} \\ 
200008316 & \highlight{100.0} & \highlight{100.0} & \highlight{100.0}	\\ 
200021644 & \highlight{100.0} & 99.63 & \highlight{100.0}\\ 
200008003 & \highlight{100.0} & \highlight{100.0} & \highlight{100.0}	\\ 
200022050 & \highlight{95.14} & 94.52 & 95.21	\\ 
200006663 & \highlight{100.0} & \highlight{100.0} &	\highlight{100.0} \\ 
200005598 & 78.29 & 93.13 &	\highlight{96.90} \\ 
200003967 & 78.46 & 98.1 & \highlight{99.79}	\\ 
100249371 & \highlight{100.0} & 99.46 &	\highlight{100.0} \\ 
200021869 & 47.13 & \highlight{69.43} & 68.58	\\ 
200014685 & \highlight{100.0} & \highlight{100.0} & \highlight{100.0}	\\ 
200021925 & 97.99 & 99.0 & \highlight{99.50} \\ 
100241706 & \highlight{100.0} & \highlight{100.0} & \highlight{100.0}	\\ 
200021763 & 32.62 & 41.18 & \highlight{84.31}	\\ 
200021637 & 8.71 & 5.65 & \highlight{98.35}	\\ 
\hline
Overall & 92.37 & 96.82 & \highlight{98.95}\\
\hline
\end{tabular}
\label{tab:accrate}
\end{table}}}

{\renewcommand{\arraystretch}{1.0}
\setlength{\tabcolsep}{16pt}{
\begin{table}[h!]
\centering
\caption{Recall Rate (over queries from length 2-20) by Book for each Model.}
\begin{tabular}{|l|c|c|c|}
\hline
Book ID& \textsc{Simple-Rules} & \textsc{Adaptive-Rules} & \textsc{Deep-AR} \\ \hline
hnsd00000 &  \highlight{99.36} &  99.11 & 99.26 \\
brsk00000 & 81.53 &  \highlight{97.55} & 96.47 \\
100249476 & 89.75 & 97.11 & \highlight{100.0} \\
200021851 & 95.9 &  \highlight{100.0} &  \highlight{100.0} \\
200018243 &  \highlight{100.0} &  \highlight{100.0} &  \highlight{100.0} \\
200021802 & 92.62 & 95.75 &  \highlight{99.2} \\
200005798 & 98.92 & 98.22 &  \highlight{100.0} \\
200010454 &  \highlight{100.0} &  \highlight{100.0} &  \highlight{100.0} \\
200021660 & 85.1 & 88.32 &  \highlight{91.49} \\
200003076 & 98.09 & 98.1 &  \highlight{98.67} \\
100249376 &  \highlight{100.0} &  \highlight{100.0} &  \highlight{100.0} \\
200004107 &  \highlight{100.0} & 96.77 &  \highlight{100.0} \\
200015779 & 77.53 & 82.15 & \highlight{96.55} \\
umgy00000 & 85.34 & 89.52 &  \highlight{96.58} \\
100249416 &  \highlight{100.0} & 96.04 &  \highlight{100.0} \\
200004148 & 84.98 & 97.19 &  \highlight{98.25} \\
100249537 &  \highlight{100.0} &  \highlight{100.0} &  \highlight{100.0} \\
200003803 &  \highlight{100.0} &  \highlight{100.0} &  \highlight{100.0} \\
200021712 &  \highlight{100.0} &  \highlight{100.0} &  \highlight{100.0} \\
200021853 & 28.12 & 32.68 &  \highlight{68.62} \\
200014740 &  99.16 & 96.58 & \highlight{99.46} \\
200008316 &  \highlight{100.0} &  \highlight{100.0} &  \highlight{100.0} \\
200021644 &  \highlight{100.0} & 96.3 & \highlight{100.0} \\
200008003 &  \highlight{100.0} &  \highlight{100.0} & \highlight{100.0} \\
200022050 & 73.7 & 72.19 &  \highlight{74.68} \\
200006663 &  \highlight{100.0} &  \highlight{100.0} &  \highlight{100.0} \\
200005598 & 68.6 & 88.51 & \highlight{90.96} \\
200003967 & 68.95 & 92.2 &  \highlight{98.44} \\
100249371 &  \highlight{100.0} & 99.65 &  \highlight{100.0} \\
200021869 & 52.15 & 54.72 &  \highlight{66.81} \\
200014685 &  \highlight{100.0} &  \highlight{100.0} & \highlight{100.0} \\
200021925 & 97.12 & 97.67 &  \highlight{97.76} \\
100241706 &  \highlight{100.0} &  \highlight{100.0} &  \highlight{100.0} \\
200021763 & 21.96 & 44.33 &  \highlight{74.96} \\
200021637 & 2.44 & 9.2 &  \highlight{90.52} \\
\hline
Overall & 89.52 & 92.69 & \highlight{96.83}\\
\hline
\end{tabular}
\label{tab:recallrate}
\end{table}}}

{\renewcommand{\arraystretch}{1.0}
\setlength{\tabcolsep}{10pt}{
\begin{table}[h!]
\centering
\caption{Recall Rate by Query Length for each Model.  Note that we only report recall since all models are designed to make prediction sequences of the same length.  }
\begin{tabular}{|l|c|c|c|}
\hline
Query Length & \textsc{Simple-Rules} & \textsc{Adaptive-Rules} & \textsc{Deep-AR} \\ \hline
1 & \highlight{100.0} & \highlight{100.0} & \highlight{100.0} \\
2 & 93.18 & 98.07 & \highlight{99.29} \\
3 & 91.98 & 96.67 & \highlight{98.75} \\
4 & 91.36 & 95.57 & \highlight{98.32} \\
5 & 90.88 & 94.72 & \highlight{97.94} \\
6 & 90.51 & 94.05 & \highlight{97.63} \\
7 & 90.19 & 93.53 & \highlight{97.37} \\
8 & 89.88 & 93.09 & \highlight{97.13} \\
9 & 89.59 & 92.7 & \highlight{96.89} \\
10 & 89.34 & 92.36 & \highlight{96.68} \\
15 & 88.27 & 90.93 & \highlight{95.87} \\
20 & 87.37 & 89.76 & \highlight{95.27} \\
25 & 86.78 & 88.74 & \highlight{94.73} \\
30 & 85.95 & 87.72  & \highlight{94.19} \\
40 & 85.06 & 86.34 & \highlight{93.83} \\
50 & 84.06 & 85.03 & \highlight{93.17} \\
\hline
\end{tabular}
\label{tab:recallquerylength}
\end{table}}}

\subsection{Combining Models}

We also find that these models tend to make different kinds of mistakes and thus if we aggregate the predictions from the different models, it is likely that at least one of them is correct, which allows us to build a higher recall combined-model.  We found that in half of the books, we're able to improve recall by aggregating predictions from all of the models together.  The overall recall is improved from 96.04\% to 97.02\%.  This aggregation of results with lower quality models comes at a sizable cost in precision (from 96.04\% to 89.44\%), so it only makes sense when we strongly prefer recall over precision.  This could occur, for example, in the case of a keyword search engine, where a user may benefit from the option of receiving a larger number of results and using their own judgement to filter out mistakes.  Additionally, this improvement in recall from aggregation provides evidence that our different models make mistakes which are somewhat independent and that they have their own unique strengths and weaknesses.  

\clearpage
\section{Conclusion}
Historical Japanese document research is a growing field, with an increasing amount of work focusing on machine learning and deep learning.  While much of the current effort has been focused on the OCR and computer vision tasks, some of the most exciting work in the future will focus on applying information retrieval and natural language understanding to these historical documents.  Our work serves as an important bridge towards successful natural language processing, by using machine learning to extract the sequential ordering of characters in historical Japanese documents.  We explored three approaches: a rule-based approach, a Adaptive-Rules-based approach, and a deep autoregressive sequence model.  The deep autoregressive model achieved the highest accuracy and requires no assumptions about layout or reading order, but has the trade-off that it requires labeled data and computational resources for training.  The results from our model are strong: 98.65\% accuracy and 97.37\% recall over 5 character queries, suggesting that it is already a feasible choice for many downstream information extraction and NLP tasks.  We also found that for books with very complex layouts (8.5\% of the books we studied), accuracy is below 80\%, suggesting that in special cases the sequence prediction task still has significant room for improvement.  





%
%

\clearpage
\bibliographystyle{splncs04}
\bibliography{refs}

\end{document}